\documentclass{ecai} 

\usepackage{latexsym}
\usepackage{amssymb}
\usepackage{amsmath}
\usepackage{amsthm}
\usepackage{booktabs}
\usepackage{enumitem}
\usepackage{graphicx}
\usepackage{color}
\usepackage{multirow}

\newcommand{\BibTeX}{B\kern-.05em{\sc i\kern-.025em b}\kern-.08em\TeX}

\begin{document}


\begin{frontmatter}


\paperid{6795} 


\title{Timing Matters: Enhancing User Experience through Temporal Prediction in Smart Homes}


\author[]{\fnms{Shrey}~\snm{Ganatra}\footnote{Equal contribution.}}
\author[]{\fnms{Spandan}~\snm{Anaokar}\footnotemark}
\author[]{\fnms{Pushpak}~\snm{Bhattacharyya}} 

\address[]{Indian Institute of Technology, Bombay}


\begin{abstract}
The proliferation of IoT devices generates vast interaction data, offering insights into user behaviour. While prior work predicts what actions users perform, the timing of these actions—critical for enabling proactive and efficient smart systems—remains relatively underexplored. Addressing this gap, we focus on predicting the time of the next user action in smart environments. Due to the lack of public datasets with fine-grained timestamps suitable for this task and associated privacy concerns, we contribute a dataset of 11.6k sequences synthesized based on human annotations of interaction patterns, pairing actions with precise timestamps. To address, we introduce \textit{Timing-Matters}, a Transformer-Encoder based method that predicts action timing, achieving 38.30\% accuracy on the synthesized dataset, outperforming the best baseline by 6\%, and showing 1--6\% improvements on other open datasets. Our code and dataset will be publicaly released.

\end{abstract}
\end{frontmatter}
%

\section{Introduction}
In the era of technological advancement, the rapid growth
of Internet of Things (IoT) solutions has resulted in an increase of smart devices within households, which is likely
to almost double from 15.1 billion in 2020 to more than 29
billion IoT devices in 2030 \cite{iotNumDevices}. This has changed
the way we interact with technology in our daily lives. From
smart thermostats to voice-activated assistants, these devices
have seamlessly integrated into our homes, workplaces, and
public spaces, providing convenience, efficiency, and connectivity like never before.

Beyond task automation and real-time data provision,
smart devices also have the potential to impact how machines understand and respond to human behaviour. These devices offer insights into our routines, lifestyle choices, and cognitive processes by continuously collecting data on user interactions, preferences, and habits. Leveraging these data can lead to the development of more intuitive and personalized technologies that anticipate the time
of our needs and preferences, ultimately enhancing our over-
all quality of life.

A crucial yet often overlooked aspect of user behaviour
prediction is the timing of actions. While existing research \cite{jeon2022accurate, xiao2023user, xiao2023know}
focuses mainly on predicting the next user action, little
attention has been given to forecasting when these actions
will occur. 
Understanding and predicting the timing of user actions is crucial, as the majority of existing research on action prediction relies on time as a contextual feature; without accurate temporal modeling, predictions risk being misaligned with real user intent and behavior.

\begin{figure*}
    \centering
    \includegraphics[scale=0.25]{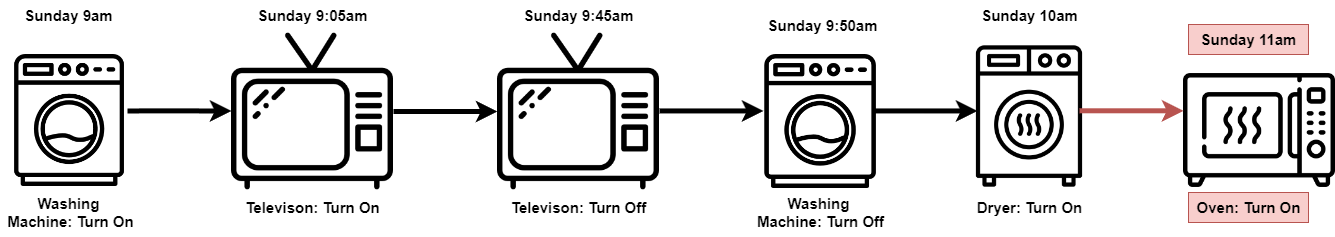}
    \caption{{Given the history of actions taken by the user we aim to predict the time of action}}
    \label{fig: Sequence 1}
\end{figure*}

Consider a smart home system designed to help with
daily routines. Suppose that the system can predict that a user will typically turn on the heater at 7:30 PM in winters. Even without applying automation to avoid energy wastage, the ability to recommend future events to users holds significant potential for smart devices and appliances.
Other example (Figure \ref{fig: Sequence 1}) shows the activity of the user on a Sunday morning. As per daily/weekly behaviour the user is likely to start preparing for lunch. Without such temporal predictions, the system remains reactive, responding merely to user commands rather than anticipating needs.

Dataset provided in SmartSense \cite{jeon2022accurate} consists of real-world user interactions but lacks exact timestamps. To protect user privacy, Time is represented into 8 3-hour bins. To overcome this, we ask human annotators to simulate their behaviour in a smart home and create a new dataset with precise timestamps.

By addressing this
novel task of action time prediction, our research takes the first steps towards creating automated assistive AI systems capable of significantly
improving the user experience in smart homes.

Our contributions are:
\begin{enumerate}
    \item \textbf{Synthesized Dataset:} We introduce a novel dataset that includes data from 16 devices and 121 device controls. The dataset contains 11,665 instances, each representing a sequence of 10 consecutive actions with precise timestamps. (Section \ref{Dataset})
    \item \textbf{Timing-Matters:} We propose a novel method, \textit{Timing-Matters}, for predicting the timing of user actions. It achieves 38.30\% accuracy on the synthesized dataset, outperforming the best baseline by approximately 6\%, and shows consistent improvements of 1--6\% across other datasets. (Section \ref{Methodology})
    \item \textbf{Context Window and Binning Analysis:} We perform an extensive analysis on the impact of context window length and number of prediction bins when modeling this prediction problem. (Section 6.5)
    \item \textbf{Analysis of Regression vs. Classification Modeling:} We conduct an in-depth study comparing the effectiveness of modeling this problem as a regression versus a classification task, highlighting the strengths and limitations of each approach. (Section 6.2)
    \end{enumerate}

\section{Related Work}
\subsection{User Behaviour Prediction in Smart Homes}
The primary focus of research on user behaviour prediction in smart homes is on learning-based techniques, specifically deep learning and conventional machine learning techniques. Traditional methods like Hidden Markov
Models (HMMs) ~\cite{lin2018sequential,sikder2019aegis} have been used to extract the user’s behaviour pattern and detect anomalies. However, HMM will fail due to the independence assumption
\cite{eddy1996hidden} of HMM which makes it unable to consider context information. \cite{alaghbari2022activities,du2019novel,farayez2018spade,tax2018human,zou2023iotbeholder} exploit Long Short-Term Memory
Networks (LSTM) to predict user behaviour in smart home scenarios. However, LSTM can only model long-term
sequential influence \cite{graveslong}, but misses out on the complex heterogeneous transitions and periodic sequential patterns
caused by users’ routines and intents. SmartSense \cite{jeon2022accurate} adopts knowledge transfer to
exploit user intent and employs a two-stage encoder to mine contextual information. However, it does not classify the routines before inputting the sequence into the encoder. Second, complex heterogeneous transitions are not modeled. Third, it does not take into account multilevel periodicity
due to a lack of consideration of historical sequences. DeepUDI \cite{xiao2023user} employs the attention mechanism to consider
the periodicity of human behaviour and the intent-aware encoder to consider intents. SmartUDI \cite{xiao2023know} considers routines, intent, and multilevel periodicity at the same time. However, none of the techniques attempts to predict the time when the user will perform the action.

\subsection{Sequential User Behaviour Prediction}
Sequential user behaviour prediction methods can be categorized as follows. Traditional methods include Factorizing
Personalized Markov Chains (FPMC) \cite{rendle2010factorizing} factorize the users-locations matrix to generate user general preferences to complete the next location prediction. However, the location independence assumption prevents FPMC from
capturing complicated sequential information \cite{rendle2010factorizing}. CNN-based methods like Caser \cite{tang2018personalized} employ CNN in
both the time-axis and feature-axis to capture temporal dynamics in sequential recommendation. However, due to the
limited receptive field of CNN, it cannot fully model long-term contextual information \cite{kim2017convolutional}. RNN-based methods like CARNN \cite{liu2016context} and SIAR \cite{rakkappan2019context} incorporate contextual information into the RNN for sequential recommendation.
However, CARNN and SIAR can not model multi-level periodicity because they only mine contextual information
in a single sequence \cite{liu2016context,rakkappan2019context}. Although DeepMove \cite{feng2018deepmove} uses attention-based RNN to model the correlation between
the current sequence and historical sequences, it leads to suboptimal performance and long prediction time since
it ambiguously considers all history sequences. SRGNN \cite{wu2019session,xu2019graph} applies gated GNN to
capture complex transition patterns among nodes for a session-based recommendation. However, SRGNN ignores
the heterogeneity of the transition patterns caused by users’ intents.
SASRec \cite{kang2018self} utilizes unidirectional transformers to capture sequential patterns in sequences while considering
the importance of correlations between behaviours. However, SASRec only captures the sequential patterns in
single sequences, which prevents it from considering multi-level periodicity.

\subsection{Time Prediction}
Time prediction—forecasting the exact time a user will perform an action—has received limited attention in smart home research. Unlike time series forecasting, which predicts values over time, or event prediction, which focuses on what will happen, time prediction requires modelling user routines, contextual patterns, and multilevel periodicity. Traditional approaches, including regression and survival analysis \cite{zhao21eventpredictionsurvey}, lack the capacity to capture sequential dependencies or dynamic contextual factors, making them unsuitable for behaviour-rich environments like smart homes. Classical point processes, such as Hawkes processes \cite{laub2015hawkesprocesses}, model inter-event dependencies but assume stationary or parametric dynamics, which do not align with the complex, periodic nature of human activity \cite{zhao21eventpredictionsurvey,benzin2023surveyeventpredictionmethods}.Neural Temporal Point Processes (NTPPs) improve flexibility by learning event time distributions via RNNs or attention mechanisms \cite{shchur2021neuraltemporalpointprocesses}. However, NTPPs primarily model inter-event times rather than absolute time, and fail to incorporate periodicity, routines, or intent—core elements in smart home behaviour. Moreover, most surveyed models ignore contextual semantics (e.g., day of week, device interactions), limiting their applicability in domestic settings \cite{benzin2023surveyeventpredictionmethods}.

In summary, while existing event time models offer valuable insights, they fall short in modelling the structured, routine-driven, and context-rich temporal patterns necessary for accurate time prediction in smart homes.

\section{Problem Statement}
This paper addresses the problem of predicting the timing of a user's next action within a smart environment, based on their sequence of preceding actions. Each historical action is represented by its device identifier, the specific control used, the day of the week, and the time it occurred. Our objective is to develop a model that leverages this sequence of past interactions to forecast *when* the subsequent action is most likely to take place, thereby providing valuable insights into temporal user behaviour patterns. For instance, given a sequence of $n-1$ user actions like the one illustrated in Fig. \ref{fig: Sequence 1}, the model's task is to predict the time of the next ($n^{th}$) action.
\\


\begin{table*}[t]
    \centering
    
    \label{tab:dataset-stats}
    \begin{tabular}{lrrrrrr}
        \hline
        \textbf{Name} & \textbf{Region} & \textbf{Time period (Y-M-D)} & \textbf{\# Instances} & \textbf{\# Devices} & \textbf{\# Device Controls} \\
        \hline
        KR & Korea & 2021-11-20 $\sim$ 2021-12-20 & 285,409 & 38 & 272 \\
        US & USA & 2022-02-22 $\sim$ 2022-03-21 & 67,882 & 40 & 268 \\
        SP & Spain & 2022-02-28 $\sim$ 2022-03-30 & 15,665 & 34 & 234 \\
        FR & France & 2022-02-27 $\sim$ 2022-03-25 & 4,423 & 33 & 222 \\
        AN & (synthesized) & 2023-04-01 $\sim$ 2023-12-31 & 11665 & 16 & 121 \\
        \hline
    \end{tabular}
    \caption{{Statistics of Device Log Datasets}}
\end{table*}
 \subsection*{TIME OF ACTION PREDICTION}
For each user instance $u$, we are given:

\begin{description}
    \item[Input] A sequence of historical actions $S_u = [x_{u,1}, \ldots, x_{u,(n-1)}]$, where each action $x_{u, i} = (d_{u, i}^{(1)}, d_{u, i}^{(2)}, c_{u, i}^{(1)}, c_{u, i}^{(2)})$ is a quadruplet representing the indices of the device, device control, day of the week, and time, respectively, for the $i^{th}$ action in the sequence.
    \item[Output] The predicted time interval $\hat{y}_u$ in which the next ($n^{th}$) action is expected to occur. Our primary approach models this as a classification task, dividing the day into $96$ discrete 15-minute intervals. While this 96-class classification formulation is the main focus of our method description, we also evaluate the impact of different time granularities  and compare the classification approach against regression-based modeling in our analysis (Section 6).
\end{description}

 \section{Methodology} 
 \label{Methodology}

\subsubsection{Time2Vec Embedding}
This is a widely used representation of time and ensures that both the sequential and cyclic patterns of time are captured \cite{kazemi2019time2vec}. This representation can be likened to a form of Fourier Transform. Except for the linear element, the rest of the elements split the time into different frequencies. This extracts the cyclic patterns present in the temporal information. For a given scalar time $\tau$, the $k$ sized time representation \textit{t2v} is given by:
$$\text{t2v}(\tau)[i] = \begin{cases} 
      \omega_i \tau + \psi_i & \text{if } i = 0 \\
      F(\omega_i \tau + \psi_i) & \text{if } 1 \leq i \leq k-1 
   \end{cases}$$
Here, \( \text{t2v}(t)[i] \) denotes the $i$th element of vector $ \text{t2v}(t) $, $ F $ is a periodic activation function with sine being the most common, and $ \omega_i $ and $ \psi_i $ are learnable parameters. 

\subsubsection{Radial Basis Functional Embedding} 
While Time2Vec incorporates the periodic properties of time, Radial Basis Functions (RBF) measure the similarity between a given scalar time and an agreed reference point. Taking inspiration from \cite{timeEncodingRBF}, we propose a new type of embedding in which each element of the vector representation denotes the Gaussian-like distance between the scalar time and a trainable reference point, For a given scalar time $\tau$, the $k$-sized time representation is given by:
$$\text{rbf}(\tau)[i] = e^{-\frac{|\tau-\mu_i|}{\sigma_i}} \quad 0 \leq i \leq k-1$$ 
where both $\mu_i$ and $\sigma_i$ are trainable parameters.\\ 
Thus we have utilised two different methods of embedding scalar time. The RBF method is dependent upon the exponential distance between the time and a certain set of trainable points in space. On the other hand, Time2Vec is a representation where the cyclic properties and emphasized. Together these two methods give complete information about the temporal properties of a sequence.

\subsection{Our Model} 

\begin{figure*}[t]
    \centering
    \includegraphics[height=110mm]{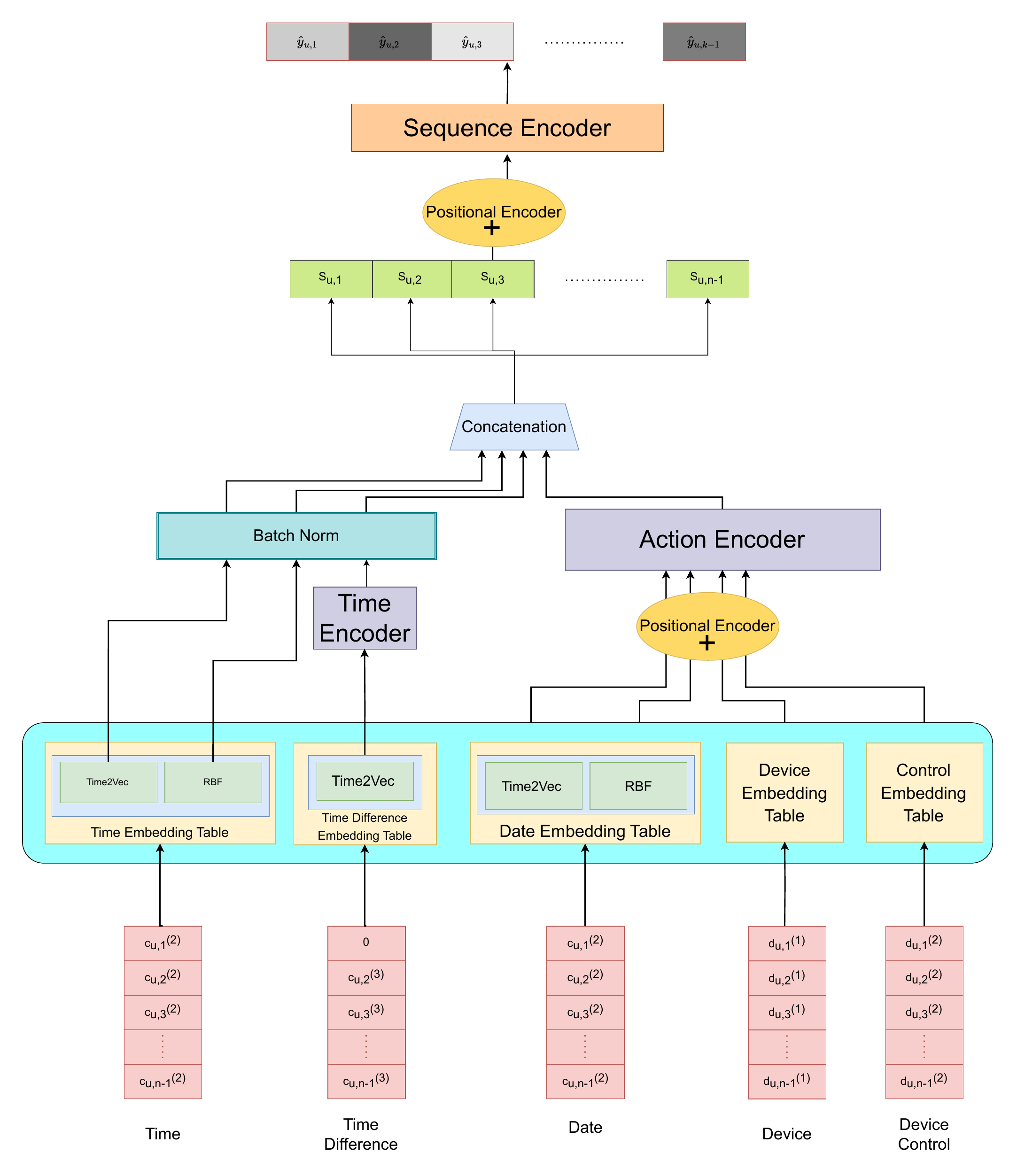}
    \caption{\textbf{The overall architecture of our model. For each session $S_u$, we send $n-1$ Action Sets, each having time $c_{u, i}^{(1)}$, date $c_{u, i}^{(2)}$, time difference $c_{u, i}^{(3)}$, device $d_{u, i}^{(1)}$ and device control $d_{u, i}^{(2)}$ to get $\hat{y}_{u, k}$ as the output which is the prediction probabilities for each action}}
    \label{fig: Model 1}
\end{figure*}

The model takes a $N \times 10 \times 4$ tensor as input and gives the predicted probabilities for all time intervals of a day. Currently, for each action, we have the time $c_{u,i}^{(1)}$, date $c_{u,i}^{(2)}$, device $d_{u,i}^{(1)}$ and device control $d_{u,i}^{(2)}$. Using this, we first derive a new feature that contains the time difference between consecutive actions $c_{u, i}^{(3)}$ that denotes the time difference between $i^{th}$ and $(i-1)^{th}$ action in a sequence. 

\subsubsection{Embedding}
Next, we rely on our embedding to convert the scalar values for different quantities into vectors. We rely on linear embedding for the device and device control $e_{u, i}^{(1)}$ and $e_{u, i}^{(2)}$. For the temporal values, we utilize both the Time2Vec and Radial Basis Function approach to get a periodic embedding and a radial embedding respectively for the time and the date the action happened at giving us $z_{u, i}^{(1, 1)}$, $z_{u, i}^{(1, 2)}$, $z_{u, i}^{(2, 1)}$, $z_{u, i}^{(2, 2)}$. The time difference, however, is converted to an embedding using Time2Vec after multiplying it by a trainable factor as we are interested in the periodic properties itself giving us $z_{u, i}^{(3)}$. \\

\subsubsection{Action Encoder}
We make use of the famous Transformer Encoder Architecture \cite{vaswani2017attention} to encode all contextual information into a single vector. This includes the device, device control, and the two temporal embeddings of the day of action. 
$$\mathbf{X} = concat\left(e_{u,i}^{(1)}, e_{u,i}^{(2)}, z_{u,i}^{(2, 1)}, z_{u,i}^{(2, 2)}\right)$$
$$\mathbf{Z} = Trans(\mathbf{X})$$
Here we further flatten the transformer encoder output,
$$\mathbf{Z'} = flatten\left( \mathbf{Z}\right)$$
$$\mathbf{H} = \mathbf{Z} \mathbf{W}$$
where $\mathbf{X} \in \mathrm{R}^{4 \times d}$, $\mathbf{Z'} \in \mathrm{R}^{4d}$, $\mathbf{H} \in \mathrm{R}^{d}$ and weight $\mathbf{W} \in \mathrm{R}^{4d \times d}$

\subsubsection{Time Encoder}
For each instance, we have $z_{u,i}^{(3)}$. First of all, we append a zero vector at the start, denoted as $z_{u,1}^{(3)}$ ensuring the sequence length is the same as that of the action encoder output. Now we pass this through a Temporal Convolution Network \cite{lea2017temporal} to get $\hat{z}_{u, i}^{(3)}$ which contains the temporal information at the neighbourhood of each element of the sequence. \\
Thus, we combine the periodic information, the radial distance, and temporal neighbourhood information which are the vectors $z_{u, i}^{(1, 1)}$, $z_{u, i}^{(1, 2)}$ and $\hat{z}_{u, i}^{(3)}$ and concatenate them along with a  Batch Normalization thereby producing a sequence of vectors in which all the temporal information for a given instance has been encoded into.

Further on, we thus concatenate to this the Action embedding getting a sequential encoding $\hat{s}_{u, i}$ of the entire sequence of action, with both the contextual and temporal elements.

\subsubsection{Sequence Encoder}
The goal of the Sequence Encoder is to finally calculate the predicted probabilities $\hat{y}$ from the final sequence $\hat{s}_{u, i}$. To do this, the encoder relies on a two-step process. \\

The first step involves using a transformer encoder and its self-attention mechanism to determine the correlation between different actions in the sequence. Due to our heavy focus on the temporal context, the correlation is also heavily temporal-dependent. Further on, we use a Positional Encoder to add positional information to the matrix, as in any regular Transformer.
$$F = Trans(S)$$
$$F' = F + P$$
where $P \in \mathrm{R}^{(t-1) \times 4d}$ is a trainable matrix. \\
The second step makes use of a Temporal Convolution Network followed by an Average Pooling Layer in 1D to get a final vector of size $4d$. This ensures that the short-term and long-term temporal dependencies are captured. We further pass it through 2 Linear layers alongside a leaky Relu activation. The size of the hidden layer changes from $4d \rightarrow 2d \rightarrow$ \textit{output class size}.

\subsubsection{Objective Function}
We rely on Cross-entropy loss to train our model. The loss function becomes:
$$L(X,Y) = -\frac{1}{n} \sum_{u} \sum_{i} y_{u,t}(i) \log \hat{y}_{u,t}(i)$$

 \section{Dataset and Experimental Setup}

We introduce our experimental setup: datasets, baselines, evaluation metrics, experimental process, and hyperparameters.

\subsection{Dataset}
 \label{Dataset}
\begin{figure*}[h]
    \centering
    \includegraphics[width=0.75\textwidth]{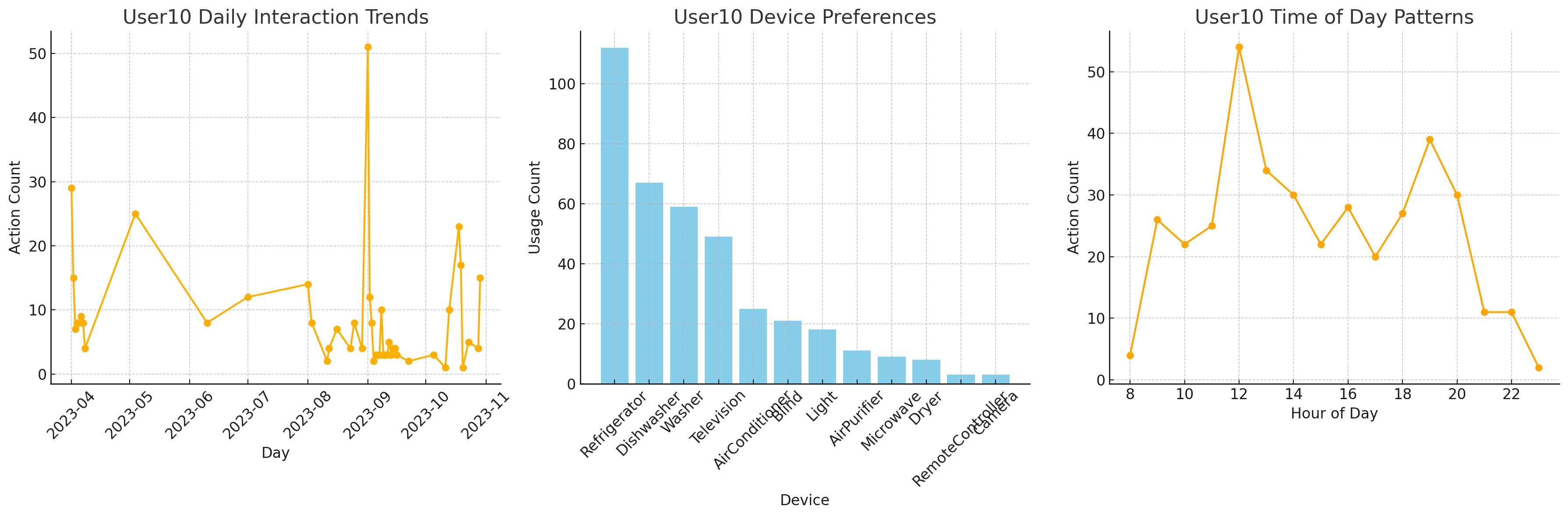}
    \caption{Analysis of AN Dataset's User10}
    \label{fig: User}
\end{figure*}

\subsubsection{SmartSense Dataset}
The dataset \cite{jeon2022accurate} contains real-world SmartThings data collected from various regions to evaluate performance. It comprises four log datasets and three routine datasets. The log datasets document device control histories executed by users of Bixby, serving as sequential instances for a general sequential recommendation. \\
The provided datasets are collected from SmartThings, a worldwide Internet of Things (IoT) platform with 62 million users. The log datasets are gathered over a month from four different countries: Korea (KR), the USA (US), Spain (SP), and France (FR). Additionally, dictionaries mapping names to IDs for various categories, such as day of weeks, hours, devices, controls, and device controls, are provided. \\
For the log dataset, each instance is represented as a tensor of size $[N \times 10 \times 5]$, where N denotes the number of instances. Each instance comprises ten consecutive actions performed by a Bixby service user, with each action represented as a vector of length 5, indicating the day of week, hour, device, control ID, and device control in order. The hour in a context is one of the 8-time ranges of 3 hours length: 0-3, 3-6, 6-9, 9-12, 12-15, 15-18, 18-21, and 21-24. The day, on the other hand, is an integer from 0 to 6. \\
\begin{table}[t]
\centering
\begin{tabular}{|l|l|l|l|l|}
\hline
\textbf{Dataset} & \textbf{Task} & \textbf{Precision(96)} & \textbf{Precision(08)} & \textbf{RMSE} \\ \hline
\multirow{2}{*}{FR} & R & 0.1445 & 0.1445 & 63978.875 \\ \cline{2-5} 
 & C & \textbf{0.3567} & \textbf{0.3567} & \textbf{30225.904} \\ \hline
\multirow{2}{*}{AN} & R & 0.2684 & \textbf{0.9329} & 8894.975 \\ \cline{2-5} 
 & C & \textbf{0.3865} & 0.7943 & \textbf{8804.339} \\ \hline
\end{tabular}
\caption{Comparison of R-Regression and C- Classification Methods}
\label{table:comparison}
\end{table}
\begin{table*}[t]
\begin{center}
\begin{tabular}{|c|c|c|c|c|c|c|c|}
\hline
\textbf{Methods} & \multicolumn{2}{c|}{\textbf{AN}} & \textbf{FR} & \textbf{SP} & \textbf{KR} & \textbf{US} \\ 
\cline{2-7}
  & \textbf{Precision(96)} & \textbf{Precision(08)} & \textbf{Precision(08)} & \textbf{Precision(08)} & \textbf{Precision(08)} & \textbf{Precision(08)} \\ 
\hline
MLP &  0.0599 &  0.4314 &  0.2891 &  0.2331 &  0.3171 &  0.2633 \\
MLP-2Step &  0.0365 &  0.4115 &  0.2448 &  0.2201 &  0.2450 &  0.2171 \\
LSTM &  0.1276 &  0.6988 &  0.2656 &  0.2220 & 0.2442 &  0.2216 \\
MLP-LSTM &  0.0156 &  0.1727 &  0.2344 &  0.2109 & 0.2439 &  0.2196 \\
LSTM-2Step &  0.0156 &  0.1727 &  0.3151 &  0.2493 & 0.2716 &  0.2820 \\
Transformer \cite{vaswani2017attention}  &  0.0182 &  0.3316 &  0.2318 &  0.1986 &  0.0867 &  0.3172 \\
BERT4Rec \cite{bert4rec} &  0.0269 &  0.2231 &  0.2839 &  0.2305 &     0.2563 & 0.3128 \\
CA-RNN \cite{carnn} &  0.1128 &  0.6658 &  \underline{0.3646} &  \underline{0.3099} &  0.4014 & \underline{0.4235} \\
SASRec \cite{kang2018self} &  0.2318 &  0.7639 &  0.3073 &  0.2311 & 0.2729	 &  0.2412 \\
SmartSense \cite{jeon2022accurate} &  \underline{0.3247} &  \underline{0.7726} &  0.3281 &  0.2832 &  \textbf{0.4093} &  0.3042 \\
\hline
\textbf{Timing-Matters} & \textbf{0.3830} & \textbf{0.8000} & \textbf{0.3750} & \textbf{0.3680} & \underline{0.4040} & \textbf{0.4440} \\ 
\hline
\end{tabular}
\end{center}
\caption{{Accuracy of our model and baselines on 5 datasets. Our model outperforms nearly all other baselines demonstrating its power in Time Prediction.}}
\label{Tab2:Results}
  \end{table*}

 \begin{table*}[t]
\begin{center}
 \begin{tabular}{|c|c|c|c|}
 \hline
    \textbf{Methods}   & \multicolumn{2}{c|}{\textbf{AN}} & \multicolumn{1}{c|}{\textbf{FR}}  \\
    \cline{2-4}
      & \textbf{Precision(96)} & \textbf{Precision(08)} & \textbf{Precision(08)}\\
   \hline
    TimingMatters & \textbf{0.383} & \textbf{0.800} & \textbf{0.375}\\
    \hline
    TimingMatters - RBF & 0.381 & 0.789 & 0.35 \\
    TimingMatters - Time Encoder & 0.335 & 0.782 & 0.368 \\
    TimingMatters - Sequence Encoder & 0.283 & 0.781 & 0.372 \\
    \hline
 \end{tabular}
\end{center}
\caption{Ablation study of our model. We report that all three core components are integral for optimal performance.}
\label{Tab3:Ablation}
  \end{table*}

\subsubsection{Anonymous (AN) Dataset}
This dataset contains synthesized raw data to mark all the daily actions a person takes in his/her smart house. It is in the form of device logs, where we have information about the action performed using a device and its timestamp. \\
This dataset is more detailed than the Smartsense Datasets as the AN dataset contains the entire time and date information for each action. The data is represented as a tensor of size $[N \times 10 \times 5]$, where N denotes the number of instances. Each instance consists of a sequence of $10$ consecutive actions performed, and each action consists of a vector of length 5, indicating the day of the year, time of day, device, user ID and device control. In this context, the time is an integer denoting the number of seconds the action was performed after the start of the day. Here the date refers to an integer denoting the day of the year a particular action was conducted on. \\
The user ID is a novel component in our dataset whereby instead of randomly generating sequences from a multitude of users as was done in the SmartSense datasets,  we make sure each generated sequence is marked by which user executes the sequence. We have 39 different users each of whom follows a different pattern when it comes to utilization of smart devices.

\subsection{Baselines}
We compare our model with the following existing methods commonly used in sequential prediction:
\begin{itemize}

    \item \textbf{MLP} Concatenates all action embeddings after flattening and passes them through 4 fully connected layers to make the final prediction.
    
    \item \textbf{MLP-2Step} Each action is first processed independently through an MLP. The resulting representations are then concatenated and passed through another MLP for prediction.
    
    \item \textbf{LSTM} Concatenates action embeddings at each timestep and feeds the sequence into a 2-layer LSTM to capture temporal dependencies.
    
    \item \textbf{MLP-LSTM} Each action is passed through an MLP before being fed into a 2-layer LSTM for temporal modeling and final prediction.
    
    \item \textbf{LSTM-2Step} Applies an LSTM individually to each action type. The resulting sequences are then passed through another LSTM for final prediction.
    
    \item \textbf{Transformer} \cite{vaswani2017attention} Concatenates action embeddings at each timestep and passes them through stacked Transformer encoder layers to capture global sequence patterns.

    \item \textbf{SASRec} \cite{kang2018self} leverages self-attention to capture long- and short-term dependencies in user action histories, efficiently identifying relevant past actions for predictions
    \item \textbf{BERT4Rec} \cite{bert4rec} Uses a bidirectional Transformer BERT architecture for recommendation
    \item \textbf{CA-RNN} \cite{carnn} considers context-dependent features uses context-specific transition matrix in RNN cell in a sequential recommendation.
    \item \textbf{SmartSense} \cite{jeon2022accurate} applies two levels of attention.
    The model also transfers commonsense knowledge from routine data to better capture user intentions. 
\end{itemize}

 \subsection{Evaluation Metrics}
 In our problem, time prediction is a classification problem where we have to make our predictions regarding what time interval the next action is likely to be taken. Hence, our final output is the list of probability values for all classes. We rely on two evaluation metrics, 96-class Accuracy (called Precision(96)) and 8-class  (called Precision(08)) which, if we divide a day into some $k$ number of classes, denote the fraction where our prediction and true values belong to the same class. In our cases, we rely on $k = 96$ and $8$ which correspond to classes of time intervals of 15 min and 3 hrs respectively.
 
\begin{figure}[t]
    \centering
    \includegraphics[width=0.45\textwidth]{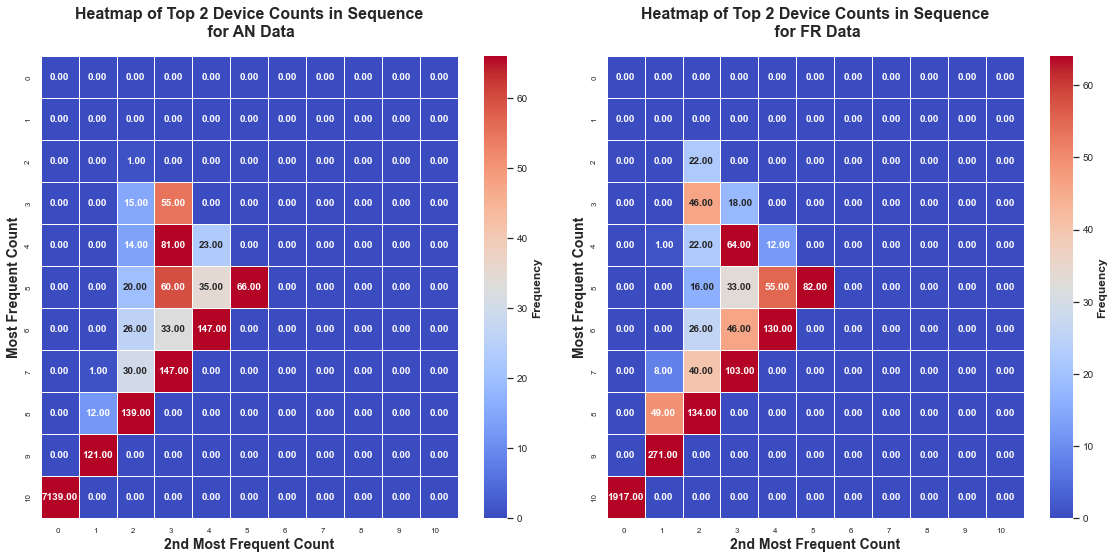}
    \caption{Heatmap of frequencies of top devices in AN and FR dataset}
    \label{fig:twofreq}
\end{figure}

 \subsection{Experimental Process}
 We first split our data into a training set, a validation set, and a test set in the ratio of $7:1:2$. For each instance in the dataset, we have 10 continuous actions. The first nine actions constitute our input, while the time of the $10^{th}$ action is our output to be predicted. \\
 Our datasets differ in the sense that the AN dataset contains information on the precise timestamp of the actions, while the Smartsense datasets contain only the range. Due to this reason, we are only able to utilize the 8-class classification on the smartsense dataset. \\
 We train the model until the validation accuracy is maximized. We then report the corresponding test accuracy.

 \subsection{HyperParameters}
 The dimension of the vector embedding is 50. All Transformer Encoders used contain multi-head structures with 2 heads and 2 layers of encoders stacked together. The feedforward network hidden layer size is 200. The Temporal Convolution Networks consist of an encoder and a decoder, where both of them contain two stacked 1D convolution units with a kernel size of 2 and a stride of 1. The first linear layer at the end of the Sequence Encoder contains a hidden layer of size 100. For training, we rely on a batch size of $64$ with a learning rate of $0.0001$ and l2reg $0.0001$.
 \section{Results and Analysis}

\subsection{Data Analysis}

\begin{figure}[t]
    \centering
    \includegraphics[width=0.45\textwidth]{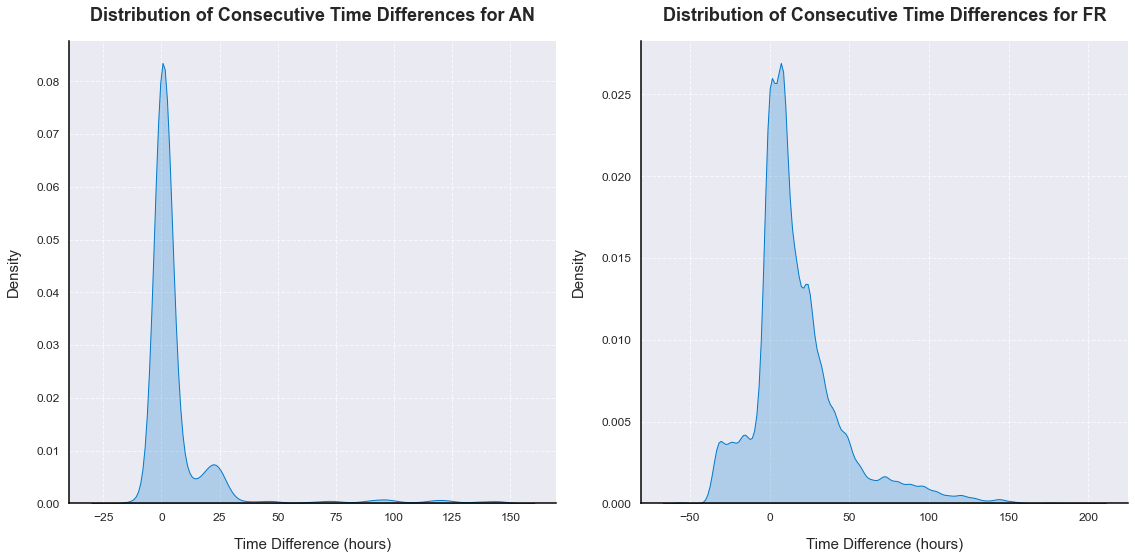}

    \caption{Consecutive Time Difference Histograms}
    \label{fig:timedifference}
\end{figure}

\subsubsection{Frequency of Actions in Sequence}
We used the SmartSense Dataset where each element is a sequence of 10 actions. Similarly, in constructing the AN dataset we also create sequences of 10 actions. In such cases, we need to understand the distribution of the actions in a given sequence and, more importantly, what devices they refer to. Taking a look at the distribution of the counts of the two most frequent devices in the sequence (Fig. \ref{fig:twofreq}), we observe that our synthetically generated dataset's actions in a sequence very closely match with that of the SmartSense Dataset.

\begin{table*}[h]
\centering

\begin{tabular}{|c|c|c|c|c|}
\hline
\multirow{2}{*}{\textbf{Context Window}} & \multicolumn{2}{c|}{\textbf{2-Layer}} & \multicolumn{2}{c|}{\textbf{4-Layer}} \\ \cline{2-5} 
 & \textbf{Precision (96)} & \textbf{Precision (08)} & \textbf{Precision (96)} & \textbf{Precision (08)} \\ \hline
5   & 0.3599 & 0.7748 & 0.3133 & 0.7578 \\ \hline
10  & \textbf{0.3830} & \textbf{0.8000} & 0.3111 & 0.7558 \\ \hline
20  & 0.2959 & 0.7641 & 0.3185 & 0.7511 \\ \hline
50  & 0.2976 & 0.7205 & 0.2786 & 0.7123 \\ \hline
100 & 0.2675 & 0.6712 & 0.1946 & 0.6887 \\ \hline
200 & 0.1819 & 0.6406 & 0.1852 & 0.6450 \\ \hline
\end{tabular}
\caption{Precision values for models with 2 and 4 transformer layers across different context windows for the AN Dataset}
\label{tab:context_size_results}
\end{table*}

\subsubsection{Time Difference between Consecutive Actions}
For the smart home dataset, unlike traditional time series datasets, each action does not occur at specific time intervals. In such cases, the consecutive time differences can actually be taken to be a time series where the $k^{th}$ element denotes the time between $k^{th}$ action and $(k+1)^{th}$. We analyze this distribution of time difference between consecutive actions. On comparing (Fig. \ref{fig:timedifference}) we find that the SmartSense dataset is more discrete, while our AN dataset provides a more continuous spectrum.

\subsection{Regression vs Classification}

Time is inherently a continuous variable, which naturally suggests treating time prediction as a regression problem. However, in our study, we found that regression is not the most effective approach for our use case. Instead, we opted to divide the day into 96 intervals, predicting which interval the next action’s time would fall into.

To evaluate the regression performance, we use the Precision(96) metric, which checks whether the predicted time falls within a 15-minute range of the true value. As shown in Table \ref{table:comparison}, the classification approach significantly outperforms regression in both Precision(96) and Test RMSE metrics. Even though the AN dataset shows better results with the coarser Precision(08) (three-hour window), the finer Precision(96) metric, offering accuracy within 15 minutes, is more effective and demonstrates the superiority of the classification method in capturing time predictions with higher precision.

 \subsection{Accuracy of Model}
 We measure the accuracy of the model and the baselines using four SmartSense Datasets and 1 Anonymous dataset. We show the AN results in terms of 96-class and 8-class classification precision. Note that the SmartSense datasets have only 8-class results, as the datasets themselves contain time information only up to a precision of 3-hour intervals. 
 

  As shown in Table \ref{Tab2:Results}, our proposed Timing-Matters model consistently outperforms the baselines across multiple datasets. For the AN dataset, our model achieves a Precision(96) of 0.383, which is a 6\% improvement compared to the best-performing baseline. In the SmartSense datasets except KR (FR, SP, and US), our model achieves the highest Precision(08) scores, with improvements ranging from 1\% to 6\% over the baselines.
 
 Overall, the results demonstrate that our model offers significant performance gains in time prediction tasks, particularly in scenarios requiring finer time granularity.

\subsection{Ablation Study}
To assess the importance of each component in our Timing-Matters model, we perform an ablation study by selectively replacing key modules and observing the impact on performance (Table \ref{Tab3:Ablation}).
\subsubsection{TimingMatters$-$RBF}
 We replace the Radial Basis Function (RBF) embedding with the Time2Vec embedding. This adjustment captures only periodic variations while ignoring radial distance information. The results show a slight drop in performance, with Precision(96) for the AN dataset decreasing from 0.383 to 0.381, and Precision(08) dropping from 0.800 to 0.789. Precision(08) decreases for FR dataset too. 
 \subsubsection{TimingMatters$-$Time Encoder} We replace the Time Encoder with an Identity function, resulting in the loss of temporal neighborhood information for each time difference element. Instead of encoding surrounding context, only the raw time difference is used. This change further reduces performance, with Precision(96) for the AN dataset decreasing to 0.335 and Precision(08) to 0.782. Precision(08) decreases slightly for FR dataset too. 
 \subsubsection{TimingMatters$-$Sequence Encoder} We replace the Transformer Encoder with an Identity function, effectively removing any correlation between consecutive actions. This modification leads to the most significant performance drop, with Precision(96) for the AN dataset decreasing to 0.283 and Precision(08) to 0.781.
 Precision(08) decreases slightly for FR dataset too. 

\subsection{Context Window} Increasing the number of historical sequences allows the model to leverage more past information, but excessive context can lead to noise and diminishing returns. As shown in Table~\ref{tab:context_size_results}, we ran Timing-Matters on the AN dataset and found using 10 historical sequences yields the best results, closely followed by 5. Due to limitations in the SmartSense dataset, we could not study the optimal context window.

\subsection{Effect of Bin Size}
The number of bins determines the temporal resolution of user activity modeling. Fewer bins simplify prediction but may miss fine-grained patterns, while more bins provide finer granularity at the cost of sparsity and performance. We studied this phenomenon by running Timing-Matters on the AN dataset as it is the only dataset with precise timestamps. As shown in Table~\ref{tab:bin_size_results}, a bin size of 96 (15-min interval) strikes an optimal balance, achieving the lowest RMSE. \textit{Precision(num\_bins)} is not used to make the choice due to its definition changing with the number of bins.

\begin{table}[t]
\centering

\label{tab:bin_size_results}
\begin{tabular}{|p{1.2cm}|c|c|c|c|c|c|}
\hline
\textbf{Number of Bins} & 8 & 12 & 24 & 48 & 96 & 288 \\
\hline
\textbf{Precision (num\_bins)} & 0.75 & 0.62 & 0.48 & 0.42 & 0.38 & 0.14 \\ \hline
\textbf{RMSE} & 12658 & 13407 & 12172 & 10896 & \textbf{10865} & 11320 \\
\hline
\end{tabular}
\caption{Effect of Bin Size on Test Precision and RMSE for the AN Dataset}
\end{table}

 \section{Conclusions and Future Work}
In this paper, we addressed the novel challenge of predicting the timing of users' next actions in smart home environments, a critical aspect for enhancing proactive AI systems. Recognizing the scarcity of suitable public datasets, we introduced a synthesized dataset with fine-grained timestamped action sequences. Our proposed model \textit{Timing-Matters} achieves the state-of-the-art performance on multiple datasets. It learns the temporal patterns very well, thus increasing its accuracy by 6\% compared to the baselines. Ablation studies confirmed the integral contribution of each model component. Furthermore, our comprehensive analyses revealed that:
(1) a context window of 10 historical actions optimally balances information gain against noise;
(2) discretizing the day into 96 15-minute intervals (bins) provides the best trade-off between temporal resolution and predictive performance; and
(3) framing the problem as a classification task significantly outperforms regression-based approaches in achieving precise timing predictions.
 

Including external factors in the model such as weather, holidays, and user calendar events can help in making more accurate predictions.



\begin{ack}
By using the \texttt{ack} environment to insert your (optional) 
acknowledgements, you can ensure that the text is suppressed whenever 
you use the \texttt{doubleblind} option. In the final version, 
acknowledgements may be included on the extra page intended for references.
\end{ack}



\bibliography{main}

\end{document}